\documentclass[11pt]{article}
\usepackage{acl}
\usepackage{times}
\usepackage{latexsym}
\usepackage{amsmath}
\usepackage{amssymb}
\usepackage{booktabs}
\usepackage{multirow}
\usepackage{graphicx}
\usepackage{subcaption}
\usepackage[T1]{fontenc}

\usepackage[utf8]{inputenc}
\usepackage{microtype}
\usepackage{inconsolata}

\usepackage{algorithm}
\usepackage{algorithmic}

\title{Awakening Dormant Experts: \\ Counterfactual Routing to Mitigate MoE Hallucinations}

\author{%
\parbox{\linewidth}{\centering
  \textbf{Wentao Hu}\textsuperscript{1,2,*,\ensuremath{\dagger}},\ 
  \textbf{Yanbo Zhai}\textsuperscript{1,*},\ 
  \textbf{Xiaohui Hu}\textsuperscript{2},\ 
  \textbf{Mingkuan Zhao}\textsuperscript{1,\ensuremath{\ddagger}},\ 
  \textbf{Shanhong Yu}\textsuperscript{4} \\
  \textbf{Xue Liu}\textsuperscript{1,2},\ 
  \textbf{Kaidong Yu}\textsuperscript{2},\ 
  \textbf{Shuangyong Song}\textsuperscript{2,\ensuremath{\ddagger}},\ 
  \textbf{Xuelong Li}\textsuperscript{3,\ensuremath{\ddagger}} \\[3pt]
  \normalfont
  \textsuperscript{1}Xi'an Jiaotong University \\
  \textsuperscript{2}Xingchen AGI Lab, China  Telecom  Artificial Intelligence  Technology (Beijing) Co., Ltd. \\
  \textsuperscript{3}Institute of Artificial Intelligence, China Telecom \quad
  \textsuperscript{4}Beijing Foreign Studies University \\[3pt]
  \small
  \texttt{\{wentao\_hu, yanbozhai, mingkuanzhao, LiuXue1087\}@stu.xjtu.edu.cn} \\
  \texttt{\{huxh12, yukd, songshy\}@chinatelecom.cn} \quad
  \texttt{yushanhong@bfsu.edu.cn} \quad
  \texttt{xuelong\_li@ieee.org}%
}}

\begin{document}
\maketitle
\renewcommand{\thefootnote}{\fnsymbol{footnote}}
\footnotetext[1]{These authors contributed equally to this work.}
\footnotetext[2]{Work done during internship at Xingchen AGI Lab.}
\footnotetext[3]{Corresponding authors.}
\renewcommand{\thefootnote}{\arabic{footnote}}
\setcounter{footnote}{0}

\begin{abstract}
Sparse Mixture-of-Experts (MoE) models have achieved remarkable scalability, 
yet they remain vulnerable to hallucinations, particularly when processing 
long-tail knowledge. We identify that this fragility stems from static 
Top-$k$ routing: routers tend to favor high-frequency patterns over rare 
factual associations. Consequently, ``specialist experts'' possessing 
critical long-tail knowledge are often assigned low gating scores and 
remain ``dormant''---under-prioritized for specific tokens despite their proven causal importance on other inputs. 
To address this, we propose Counterfactual Routing (CoR), a training-free 
inference framework designed to awaken these dormant experts. CoR integrates 
layer-wise perturbation analysis with the Counterfactual Expert Impact (CEI) 
metric to dynamically shift computational resources from syntax-dominant to 
knowledge-intensive layers while maintaining a constant total activation 
count, effectively retrieving causally decisive experts via virtual ablation. 
Extensive experiments on TruthfulQA, FACTOR, and TriviaQA demonstrate that CoR 
improves factual accuracy by 3.1\% on average without increasing the 
inference budget, establishing a superior Pareto frontier compared to 
static scaling strategies. Code is available at \url{https://github.com/ZhaiYanbo/CoR}.
\end{abstract}


\section{Introduction}\label{sec:intro}

Mixture-of-Experts (MoE) architectures have become the dominant paradigm for scaling Large Language Models (LLMs) by decoupling parameter count from inference cost~\citep{shazeer2017outrageously}. By routing each token to only a subset of experts, MoE models scale to hundreds of billions of parameters while maintaining computational efficiency comparable to much smaller dense models. For instance, Qwen-3-30B-A3B~\citep{yang2025qwen3technicalreport} activates only 3B of its 30B parameters per token, outperforming dense counterparts with significantly less computation. Similarly, the TeleChat model family has progressed from dense 
architectures~\citep{he2024telechattechnicalreport, wang-etal-2024-telechat, 
wang2025technicalreporttelechat2telechat25, li2024teleflmtechnicalreport, 
li202452b1tlessonslearned} to the TeleChat3-MoE 
series~\citep{liu2025trainingreporttelechat3moe}, scaling to over one trillion 
parameters with a high-sparsity MoE architecture, further demonstrating 
the trend of MoE as the dominant scaling paradigm. Despite these efficiency advantages, MoE models remain susceptible to hallucinations---generating plausible but factually incorrect content~\citep{ji2023survey, huang2025survey}---particularly for long-tail entities where accuracy degrades sharply~\citep{kandpal2023large, mallen2023not}.

Recent studies have identified two factors particularly relevant to MoE architectures. First, long-tail knowledge sparsity: LLMs primarily capture frequent patterns, struggling with rare facts due to insufficient training signal~\citep{sun2024head, kandpal2023large}. \citet{kalai2024calibrated} theoretically proved that hallucinations are inevitable when models generalize beyond training coverage. Second, spurious correlations: models learn misleading pattern associations that produce plausible-sounding but factually incorrect content when queried about unfamiliar entities~\citep{seitzer2022pitfalls, zhang2024co}.

These issues manifest distinctively in MoE models through the routing mechanism. Standard Top-$k$ routing is trained jointly with expert parameters using auxiliary load balancing losses~\citep{fedus2022switch, lepikhin2020gshard, zhou2022mixture}. However, since load balancing encourages uniform expert utilization and high-frequency tokens dominate the training corpus, routers learn to favor frequency-based patterns over rare factual associations. This creates a systematic correlation-causality misalignment: ``generalist experts'' handling common linguistic features receive high gating scores, while ``specialist experts'' harboring long-tail knowledge are assigned low scores. During inference on hard tokens, these specialists become ``dormant'': functionally capable of providing correct information but receiving low gating scores for the current context, despite proving critical on other inputs. The model ``knows'' the fact (stored in specialist parameters) but fails to ``recall'' it (due to routing decisions), resulting in hallucinations.

Existing hallucination mitigation techniques fail to address this routing bottleneck. Training-time approaches such as retrieval-augmented generation~\citep{lewis2020retrieval} require extensive resources and cannot be applied to deployed models. Inference-time interventions like DoLa~\citep{chuang2023dola} and ITI~\citep{li2023inference} operate on the output distribution or residual stream, attempting corrections after routing decisions have been made and cannot recover information from experts that were never activated.

To address this, we propose Counterfactual Routing (CoR), a training-free inference framework that awakens dormant experts through causal guidance. Our key insight is to distinguish correlation (what the router prefers based on training statistics) from causality (what the prediction actually needs for factual correctness). CoR achieves factuality improvement via compute-preserving expert redistribution---activating the same total number of experts as standard inference---through two complementary mechanisms. At the layer-wise level, recognizing that factual knowledge concentrates in specific layers~\citep{meng2022locating, geva2023dissecting, dai2022knowledge}, we employ perturbation analysis to identify knowledge-intensive layers and dynamically expand their expert budget while reducing allocations to syntax-dominant layers. At the expert-wise level, we introduce Counterfactual Expert Impact (CEI), a causal metric derived from virtual ablation. Unlike correlation-based gating scores, CEI captures causal necessity---identifying experts essential for factual correctness regardless of their router scores. Experts with high CEI but low router scores are precisely the dormant specialists that CoR awakens. Extensive experiments on Qwen-3-30B-A3B~\citep{yang2025qwen3technicalreport}, DeepSeek-V2-Lite~\citep{deepseekai2024deepseekv2strongeconomicalefficient}, and GPT-OSS-20B~\citep{openai2025gptoss120bgptoss20bmodel} demonstrate the superiority of our method.

Our contributions are summarized as follows:
\begin{itemize}
    \item We expose the ``Dormant Expert'' phenomenon in MoE models, providing empirical evidence showing how static Top-$k$ routing contextually under-prioritizes knowledge-bearing experts for long-tail tokens due to correlation-causality misalignment.
    \item We propose CoR, a training-free framework achieving compute-preserving expert redistribution through causal-guided resource reallocation across layers and experts.
    \item Extensive experiments demonstrate that CoR achieves an average improvement of 3.1\% on hallucination benchmarks without increasing the inference budget, establishing a superior Pareto frontier compared to static scaling strategies.
\end{itemize}

\begin{figure*}[t]
    \centering
    \includegraphics[width=0.9\linewidth]{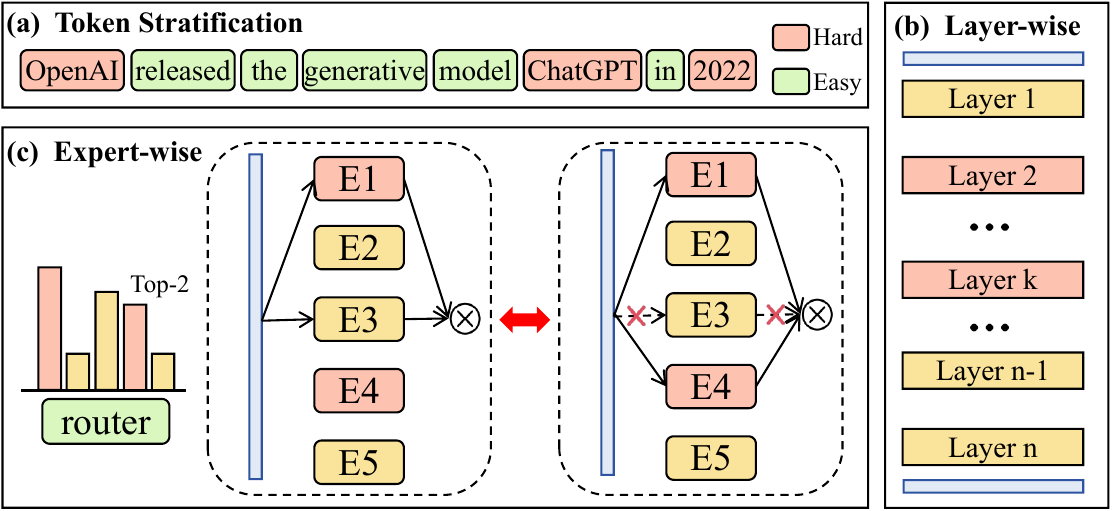}
    \caption{Overview of the Offline Causal Analysis pipeline. The framework consists of three hierarchical stages: (a) Token Stratification: We stratify tokens into hard (knowledge-intensive) and easy (syntax-dominant) subsets based on model uncertainty to disentangle factual reasoning from generic processing. (b) Layer-wise Analysis: We apply Contrastive Sensitivity Normalization to identify knowledge-intensive layers by measuring their relative sensitivity ($R_l$) to hard tokens while mitigating error cascading. (c) Expert-wise Analysis: We compute CEI via virtual ablation to uncover ``dormant'' experts---causally critical on some tokens but under-prioritized on others.}
    \label{fig:offline_pipeline}
\end{figure*}

\section{Related Work}

\paragraph{Hallucination Mitigation.}
Hallucinations in LLMs---generating plausible but factually incorrect content---have been extensively studied~\citep{ji2023survey, huang2025survey, zhang2025sirenssongaiocean}. Mitigation approaches span training-time methods like retrieval-augmented generation~\citep{lewis2020retrieval} and factuality-enhanced pretraining~\citep{lee2022factuality}, as well as inference-time interventions. DoLa~\citep{chuang2023dola} contrasts logits from different layers to amplify factual signals, while ITI~\citep{li2023inference} shifts activations along truthfulness directions. However, these techniques operate on output distributions or residual streams, essentially polishing the result after the routing decision is made. Notably, existing mitigation methods have been exclusively designed and validated on dense architectures. To our knowledge, no prior work has specifically addressed hallucinations arising from the \textit{routing-level} decisions unique to MoE mechanisms. Our work bridges this gap by intervening directly in the expert selection process. Orthogonally, recent studies have explored enhancing LLM reasoning 
capabilities through structured thinking paradigms, including computation 
logic graphs for mathematical reasoning~\citep{zhao2025enhancing}, 
multi-perspective reasoning with reinforcement learning for information 
extraction~\citep{li2025mruiemultiperspectivereasoningreinforcement}, 
table reasoning frameworks leveraging schema-guided 
decomposition~\citep{xiong2025tablereasoneradvancingtablereasoning}, 
structured reasoning data construction 
pipelines~\citep{xing-etal-2025-llmsr}, preference-driven 
methodologies for efficient code 
generation~\citep{li2025preference}, and reinforcement 
learning-driven optimization strategies for domain-specific 
tasks~\citep{Li2025Efficient}. While these methods improve 
reasoning quality at the task level, they do not address the 
routing-level decisions within MoE architectures that our work targets.

\paragraph{MoE Routing Mechanisms.}
Sparse MoE models replace dense feedforward layers with expert modules, using learned gating functions to select a subset of experts per token ~\citep{shazeer2017outrageously, fedus2022switch}. Training requires auxiliary losses for load balancing, as routers otherwise collapse to repeatedly selecting the same experts ~\citep{lepikhin2020gshard}. Expert Choice routing inverts selection, having experts choose tokens ~\citep{zhou2022mixture}. Recent analyses reveal that experts exhibit temporal locality and syntactic rather than semantic specialization ~\citep{jiang2024mixtral}. These findings suggest routing learns surface patterns rather than factual associations, motivating our causally-grounded expert selection. Despite extensive work on MoE routing efficiency and load balancing, the connection between routing decisions and factual accuracy remains underexplored. Beyond routing design, complementary directions for improving MoE 
efficiency have been explored. \citet{hu2025mosaicpruninghierarchicalframework} 
propose Mosaic Pruning, a hierarchical framework that preserves 
functionally diverse experts during structured pruning via a 
``cluster-then-select'' strategy, revealing that expert specialization 
patterns are critical for downstream generalization. At the attention 
level, \citet{zhao2025makingheadcountsparse} introduce SPAttention, 
which partitions the attention workload into non-overlapping distance 
bands across heads, demonstrating that principled structural sparsity 
can serve as an effective inductive bias without sacrificing performance. Additionally, knowledge distillation has emerged as a complementary 
paradigm for compressing large-scale models while preserving 
representational capacity~\citep{li2025frequency, li2025ammkd, 
li2026distilling}, and similarity-guided layer pruning strategies 
further reduce redundancy in deep 
architectures~\citep{li2024sglp, li2026sepprune}. Our work bridges this gap by revealing how standard routing contextually disadvantages knowledge retrieval, and proposes the first causally-grounded intervention specifically targeting this failure mode.

\section{Methodology}

\begin{figure*}[t]
    \centering
    \includegraphics[width=\linewidth]{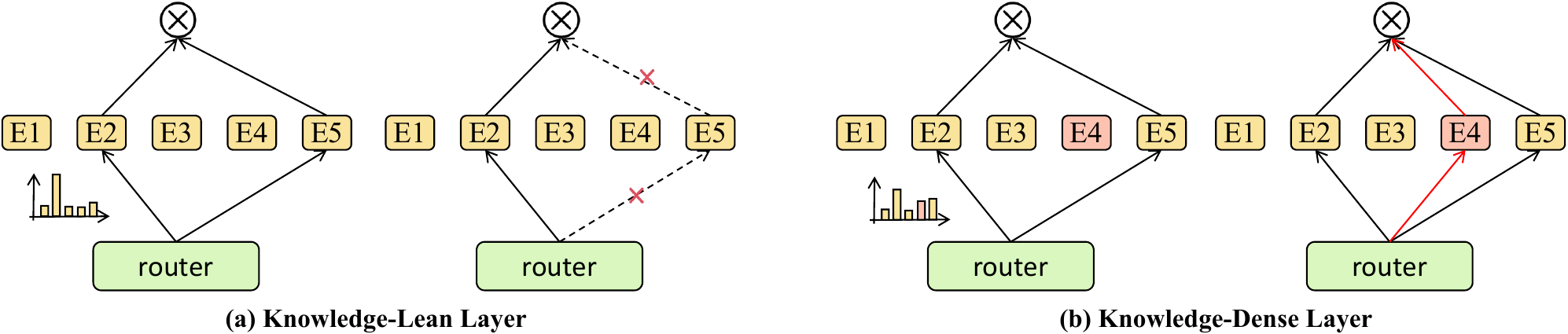}
    \caption{Schematic of Compute-Preserving Expert Redistribution. CoR dynamically reallocates budget based on layer sensitivity. (a) Knowledge-Lean Layer: reduce active experts to save budget. (b) Knowledge-Dense Layer: expand budget and fuse CEI to awaken dormant experts. The total activation count remains constant (\textit{budget-neutral}).}
    \label{fig:inference_process}
\end{figure*}

In this section, we present the Counterfactual Routing (CoR) framework. We first formalize the standard Sparse Mixture-of-Experts (MoE) architecture. We then detail our two-phase approach: an offline causal analysis to identify critical components (Figure \ref{fig:offline_pipeline}), and an online inference strategy based on compute-preserving expert redistribution (Figure \ref{fig:inference_process}).

\subsection{Preliminaries}

We consider a Transformer-based MoE model with $L$ layers. Let $\mathbf{h}_{l-1}^{(t)} \in \mathbb{R}^d$ denote the hidden state of the $t$-th token at layer $l-1$. The $l$-th MoE layer consists of $N$ expert networks $\{E_{l,i}\}_{i=1}^N$, where each expert is typically a Feed-Forward Network (FFN).

\textbf{Gating Mechanism.} A router (or gating network) $G_l$ determines the contribution of each expert. It typically applies a linear projection followed by a Softmax function:
\begin{equation}
    G_l(\mathbf{h}_{l-1}^{(t)}) = \text{Softmax}(\mathbf{h}_{l-1}^{(t)} \mathbf{W}_g)
\end{equation}
where $\mathbf{W}_g \in \mathbb{R}^{d \times N}$ is the learnable routing weight matrix.
Let $g_{l,i}^{(t)}$ denote the gating probability for the $i$-th expert. To ensure sparsity and computational efficiency, standard MoE employs a Top-$k$ selection strategy. The set of active indices for token $t$ is defined as $\mathcal{A}_l^{(t)} = \text{Top-}k ( \{g_{l,i}^{(t)}\}_{i=1}^N )$, where $k \ll N$.

\textbf{Sparse Output.} The output of the MoE layer, $\mathbf{h}_l^{(t)}$, is the weighted sum of the activated experts' outputs plus the residual connection:
\begin{equation}
    \mathbf{h}_l^{(t)} = \mathbf{h}_{l-1}^{(t)} + \sum_{i \in \mathcal{A}_l^{(t)}} g_{l,i}^{(t)} \cdot E_{l,i}(\mathbf{h}_{l-1}^{(t)})
\end{equation}

\subsection{Offline Causal Analysis}
\label{sec:offline_analysis}

Standard routing mechanisms often fail to capture long-tail knowledge due to spurious correlations learned during training. To rectify this, we perform a comprehensive offline analysis on a calibration dataset to pinpoint where factual knowledge resides (Layer-wise analysis) and which experts are causally necessary (Expert-wise analysis).

\subsubsection{Token Stratification}
To disentangle knowledge-intensive reasoning from generic syntactic processing, we stratify tokens based on model uncertainty. By performing a forward pass on a calibration dataset (e.g., C4~\citep{dodge2021documentinglargewebtextcorpora}) using the Original model $\mathcal{M}$, we compute the token-level negative log-likelihood (NLL) loss $\ell^{(t)}$. We define two contrastive subsets based on percentile thresholds: the hard sample set $\mathcal{D}_{\text{hard}} = \{ t \mid \ell^{(t)} > \tau_{\text{high}} \}$ representing long-tail knowledge, and the easy sample set $\mathcal{D}_{\text{easy}} = \{ t \mid \ell^{(t)} < \tau_{\text{low}} \}$ representing common syntax.

\subsubsection{Layer-wise analysis}
Standard MoEs allocate a uniform computational budget across all layers. However, we hypothesize that factual knowledge is concentrated in specific deep layers. To locate these layers, we employ Layer-wise Perturbation Analysis.

Crucially, raw perturbation scores are often biased by the ``Error Cascading'' effect: a small perturbation in shallow layers undergoes successive amplifications through the network's depth (via matrix multiplications and non-linearities), resulting in a spuriously high loss deviation even for semantically insensitive layers. To eliminate this structural bias and isolate true knowledge dependency, we propose Contrastive Sensitivity Normalization.

\textbf{Raw Sensitivity Calculation.} For a specific layer $l$, we apply a uniform multiplicative perturbation $\delta$ to its output $\mathbf{O}_{l}^{(t)}$ while keeping other layers frozen. We calculate the expected loss degradation for both hard and easy subsets separately:
\begin{equation}
    S_l(\mathcal{D}) = \frac{1}{|\mathcal{D}|} \sum_{t \in \mathcal{D}} \left[ \ell(\mathcal{M}(\tilde{\mathbf{O}}_{l}^{(t)})) - \ell^{(t)} \right]
\end{equation}
where $\tilde{\mathbf{O}}_{l}^{(t)} = (1 + \delta) \odot \mathbf{O}_{l}^{(t)}$ and $\mathcal{D} \in \{\mathcal{D}_{\text{hard}}, \mathcal{D}_{\text{easy}}\}$. Intuitively, $S_l(\mathcal{D}_{\text{easy}})$ captures the layer's inherent structural amplification factor, as easy tokens primarily rely on syntax processing rather than deep knowledge retrieval.

\textbf{Relative Knowledge Intensity (RKI).} To decouple the knowledge component from structural noise, we define the final layer score $R_l$ as the ratio of hard-to-easy sensitivity:
\begin{equation}
    R_l = \frac{S_l(\mathcal{D}_{\text{hard}})}{S_l(\mathcal{D}_{\text{easy}}) + \epsilon}
\end{equation}
A high $R_l$ indicates that layer $l$ is disproportionately critical for hard tokens compared to easy ones, signifying a Knowledge-Intensive Layer. This score will guide our Layer-wise analysis budget reallocation. We provide a rigorous theoretical derivation demonstrating how $R_l$ cancels out structural depth bias in Appendix~\ref{app:theory}.

\subsubsection{Expert-wise analysis}

Within knowledge-intensive layers, the router may still assign low gating scores to certain experts for specific input tokens, even though these experts prove critical when activated on other hard tokens. We term such contextually suppressed experts as ``dormant'' and propose Counterfactual Expert Impact (CEI), a causal metric that quantifies an expert's aggregated causal value across the calibration set via virtual ablation.

Specifically, for a target expert $e$ in layer $l$, we first isolate the subset of hard tokens where this expert was originally activated by the Top-$k$ mechanism, denoted as $\mathcal{T}_{l,e} = \{ t \in \mathcal{D}_{\text{hard}} \mid e \in \mathcal{A}_l^{(t)} \}$. To quantify the causal contribution of expert $e$, we construct a locally counterfactual scenario by explicitly ablating it from the network. To ensure a valid probability distribution during this intervention, the original gating probability mass of expert $e$ is redistributed proportionally among the remaining experts to form a surrogate distribution $\tilde{g}$:
\begin{equation}
    \tilde{g}_{l,j}^{(t)} = 
    \begin{cases} 
    0 & \text{if } j = e \\
    \frac{g_{l,j}^{(t)}}{\sum_{m \neq e} g_{l,m}^{(t)}} & \text{if } j \neq e 
    \end{cases}
\end{equation}
We then compute the counterfactual loss $\tilde{\ell}_{(l,-e)}^{(t)}$ using this ablated routing configuration.

The marginal causal effect of expert $e$ on token $t$ is defined as the performance degradation (loss increase) caused by its removal, termed the ``Rescue Gain'':
\begin{equation}
    \Delta \ell_{l,e}^{(t)} = \tilde{\ell}_{(l,-e)}^{(t)} - \ell^{(t)}
\end{equation}
A positive $\Delta \ell_{l,e}^{(t)}$ implies that expert $e$ possesses unique knowledge required for token $t$ that cannot be compensated by other generalist experts. Finally, the global CEI score for expert $e$ is derived by aggregating the expected causal effect across all relevant hard tokens:
\begin{equation}
    \text{CEI}_{l,e} = \mathbb{E}_{t \in \mathcal{T}_{l,e}} \left[ \Delta \ell_{l,e}^{(t)} \right]
\end{equation}
This metric effectively disentangles an expert's necessity from its popularity (gating frequency), highlighting experts that are critical for correctness despite low routing confidence. Since CEI aggregates causal impact across calibration tokens, an expert with high CEI may still receive a low router score on a \textit{new} input due to context shift---precisely the ``dormant'' scenario where CoR intervenes.

\subsection{Compute-Preserving Inference}
\label{sec:inference}

During the inference phase, we dynamically adjust the routing strategy based on the pre-computed layer sensitivity $\{R_l\}$ and expert criticality $\{\text{CEI}_{l,e}\}$. Our approach operates on the principle of compute-preserving expert redistribution, ensuring that the total computational cost remains constant while maximizing factual accuracy. The inference process is depicted in Figure \ref{fig:inference_process}.

\textbf{Adaptive Budgeting via Reallocation.} 
Instead of a uniform Top-$k$ across all layers, we reallocate the total computational budget $K_{\text{total}} = L \times k_{\text{baseline}}$ across layers proportional to their Relative Knowledge Intensity $R_l$. The number of active experts $k_l$ for layer $l$ is calculated as:
\begin{equation}
    k_l = \text{Round} \left( K_{\text{total}} \cdot \frac{R_l}{\sum_{j=1}^L R_j} \right)
\end{equation}
This strategy allocates more experts to deep, knowledge-intensive layers to enhance reasoning capabilities, while pruning activations in shallow, syntax-dominant layers.\footnote{Here, $k$ denotes the fixed number of active experts employed in the standard Top-$k$ routing baseline.}  Crucially, the constraint $\sum k_l \approx K_{\text{total}}$ ensures that the overall FLOPs count remains equivalent to the standard baseline.

\begin{figure}[h] 
    \centering
    \includegraphics[width=0.8\columnwidth]{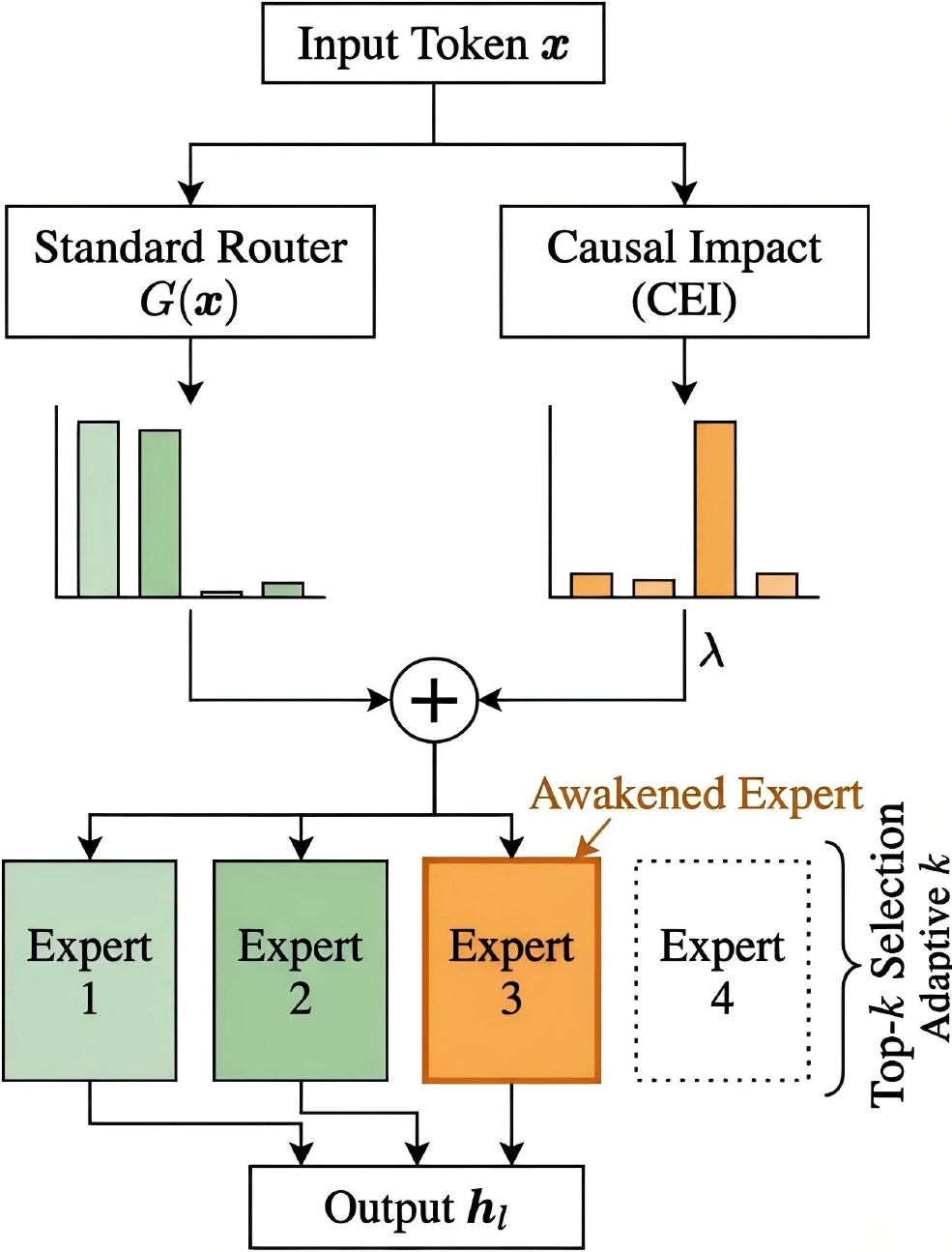} 
    \caption{Illustration of Causal-Guided Expert Awakening.
    This diagram visualizes the fusion mechanism in Eq. \ref{eq:score_fusion}. The \textit{Standard Router} (left) suppresses the critical expert (Expert 3) due to frequency bias. CoR injects the \textit{Causal Impact} (right) as a prior, boosting the fused score to ``awaken'' the dormant specialist.}
    \label{fig:fusion_mechanism}
\end{figure}

\textbf{Awakening Routing via Context-Prior Fusion.} 
Once the layer-wise budget $k_l$ is determined, we rely on a hybrid mechanism to select specific experts. Relying solely on the static CEI score would ignore the dynamic context of the current input token (e.g., syntax structure), potentially harming general fluency. Conversely, relying solely on the original router $G_l(x)$ leads to the dormant expert problem.
Therefore, we select the Top-$k_l$ experts by fusing the \textit{dynamic context} with the \textit{causal prior}, as illustrated in Figure~\ref{fig:fusion_mechanism}. The final selection score for expert $i$ in layer $l$ is defined as:

\begin{equation}
    \text{Score}_{l,i}(x) = \underbrace{G_l(x)_i}_{\text{Context}} + \lambda \cdot \underbrace{\text{CEI}_{l,i}}_{\text{Prior}}
    \label{eq:score_fusion}
\end{equation}
where $\lambda$ is a hyperparameter controlling the intervention strength. Note that to align the scales, we apply Min-Max Normalization to the raw $\text{CEI}_{l}$ scores within each layer before fusion. This formulation acts as a causal rectification: for general tokens, the high magnitude of $G_l(x)$ dominates, preserving the model's linguistic capabilities; for hard knowledge tokens where the router is uncertain, the high $\text{CEI}$ acts as a bias term to ``awaken'' the specialist expert.

\begin{table*}[t]
\centering
\footnotesize
\renewcommand{\arraystretch}{1} 
\setlength{\tabcolsep}{9pt}      

\resizebox{\textwidth}{!}{%
\begin{tabular}{cl ccc cc cc}
\toprule
\multirow{2}{*}{\textbf{Model}} & \multicolumn{1}{c}{\multirow{2}{*}{\textbf{Method}}} & \multicolumn{3}{c}{\textbf{TruthfulQA}} & \multicolumn{2}{c}{\textbf{FACTOR}} & \multirow{2}{*}{\textbf{TriviaQA}} & \multirow{2}{*}{\textbf{Average}} \\
\cmidrule(lr){3-5} \cmidrule(lr){6-7}
 & & MC1 & MC2 & Gen & News & Wiki & & \\ 
\midrule
 
\multirow{5}{*}{\textbf{Qwen-3-30B-A3B}} 
 & Standard & 34.15 & 53.27 & 40.64 & 65.35 & 56.15 & 38.49 & 48.01 \\
 & Random   & 30.42 & 49.85 & 36.20 & 61.20 & 52.30 & 34.15 & 44.02 \\
 & DoLa     & 34.38 & 53.45 & 40.92 & 65.60 & 56.10 & 38.55 & 48.17 \\
 & ITI      & 34.22 & 53.15 & 40.50 & 65.10 & 56.25 & 38.38 & 47.93 \\ 
 & \textbf{CoR (Ours)} & \textbf{35.13} & \textbf{54.81} & \textbf{44.08} & \textbf{67.80} & \textbf{58.15} & \textbf{39.70} & \textbf{49.95} \\ 
\midrule

\multirow{5}{*}{\textbf{DeepSeek-V2-Lite}} 
 & Standard & 21.54 & 31.88 & 29.13 & 56.21 & 47.73 & \textbf{42.25} & 38.12 \\
 & Random   & 18.30 & 28.50 & 25.40 & 52.15 & 44.20 & 38.60 & 34.53 \\
 & DoLa     & 21.10 & 32.05 & 28.95 & 55.90 & 47.50 & 41.80 & 37.88 \\
 & ITI      & 21.65 & 32.15 & 29.20 & 56.35 & 47.60 & 41.75 & 38.12 \\ 
 & \textbf{CoR (Ours)} & \textbf{23.26} & \textbf{43.61} & \textbf{45.90} & \textbf{58.01} & \textbf{48.73} & 41.89 & \textbf{43.57} \\ 
\midrule

\multirow{5}{*}{\textbf{GPT-OSS-20B}} 
 & Standard & 34.03 & 53.11 & 19.20 & 30.69 & 34.87 & 29.55 & 33.58 \\
 & Random   & 31.50 & 50.20 & 16.80 & 28.10 & 32.40 & 26.80 & 30.97 \\
 & DoLa     & 34.60 & 53.55 & 19.85 & 31.05 & 34.62 & 30.15 & 33.97 \\
 & ITI      & 34.45 & 53.40 & 20.10 & 30.85 & 34.95 & 29.90 & 33.94 \\ 
 & \textbf{CoR (Ours)} & \textbf{36.32} & \textbf{54.73} & \textbf{21.55} & \textbf{31.82} & \textbf{35.32} & \textbf{33.21} & \textbf{35.49} \\ 
\bottomrule
\end{tabular}%
}
\caption{Zero-shot performance of Counterfactual Routing (CoR) compared with baseline methods on factuality benchmarks. CoR is compared with Standard Top-$k$ routing, Random routing, and inference-time interventions (DoLa, ITI) across three architectures. Best scores are in bold.}
\label{tab:main_results}
\end{table*}

\begin{table*}[t]
\centering
\footnotesize 
\renewcommand{\arraystretch}{1} 
\setlength{\tabcolsep}{10pt}

\resizebox{\textwidth}{!}{%
\begin{tabular}{cl ccc cc cc}
\toprule
\multirow{2}{*}{\textbf{Model}} & \multicolumn{1}{c}{\multirow{2}{*}{\textbf{Method}}} & \multicolumn{3}{c}{\textbf{TruthfulQA}} & \multicolumn{2}{c}{\textbf{FACTOR}} & \multirow{2}{*}{\textbf{TriviaQA}} & \multirow{2}{*}{\textbf{Average}} \\
\cmidrule(lr){3-5} \cmidrule(lr){6-7}
 & & MC1 & MC2 & Gen & News & Wiki & & \\ 
\midrule
 
\multirow{4}{*}{\textbf{Qwen-3-30B-A3B}} 
 & Standard         & 34.15 & 53.27 & 40.64 & 65.35 & 56.15 & 38.49 & 48.01 \\
 & Layer-wise  & 34.35 & 53.58 & 41.42 & 65.80 & 56.55 & 38.82 & 48.42 \\
 & Expert-wise & 34.78 & 54.25 & 42.90 & 66.95 & 57.40 & 39.35 & 49.27 \\
 & \textbf{CoR (Full)} & \textbf{35.13} & \textbf{54.81} & \textbf{44.08} & \textbf{67.80} & \textbf{58.15} & \textbf{39.70} & \textbf{49.95} \\ 
\bottomrule
\end{tabular}%
}
\caption{Zero-shot ablation study of CoR components on the Qwen-3-30B-A3B. We compare the full CoR framework against the Standard baseline and variants using only Layer-wise Budget Reallocation or Expert-wise Causal Awakening. Best scores are in bold.}
\label{tab:ablation}
\end{table*}

\paragraph{Complexity and Robustness Analysis.}
We briefly analyze the overhead and hyperparameter sensitivity of our framework. In terms of offline efficiency, calculating CEI involves virtual ablation, which is a computationally intensive but \textit{one-off offline process} performed on a small calibration subset (e.g., $|\mathcal{D}| \approx 1,000$ tokens). Crucially, this pre-computation incurs zero additional latency during online inference since the causal scores are cached. Regarding hyperparameter stability, for the intervention strength $\lambda$ in Eq. \ref{eq:score_fusion}, we adopt a conservative fixed value $\lambda = 0.1$ across all experiments. This choice functions as a minimal intervention strategy: it ensures that the causal prior acts as a subtle guide to ``tip the scale'' for dormant experts only when the original router is uncertain, thereby preserving the model's general linguistic fluency without aggressive overriding. Our analysis confirms that performance is robust within the small-value regime ($\lambda \in [0.05, 0.2]$).

\begin{table*}[t]
\centering
\footnotesize 
\renewcommand{\arraystretch}{1} 
\setlength{\tabcolsep}{6pt}       

\resizebox{\textwidth}{!}{%
\begin{tabular}{cl cc cccc cc}
\toprule
\multirow{2}{*}{\textbf{Model}} & \multicolumn{1}{c}{\multirow{2}{*}{\textbf{Method}}} & \multirow{2}{*}{\textbf{ARC-C}} & \multirow{2}{*}{\textbf{ARC-E}} & \multicolumn{4}{c}{\textbf{MMLU}} & \multirow{2}{*}{\textbf{GSM8K}} & \multirow{2}{*}{\textbf{Average}} \\
\cmidrule(lr){5-8}
 & & & & Human & Social & STEM & Others & & \\ 
\midrule

\multirow{3}{*}{\textbf{Qwen-3-30B-A3B}} 
 & Standard & 52.73 & \textbf{79.88} & \textbf{67.69} & 87.49 & \textbf{78.91} & \textbf{81.69} & 86.43 & 76.40 \\
 & Random   & 45.15 & 72.52 & 60.25 & 78.61 & 70.52 & 73.46 & 75.20 & 67.96 \\
 & \textbf{CoR} & \textbf{53.24} & 79.48 & 67.44 & \textbf{87.55} & 78.88 & 81.30 & \textbf{87.23} & \textbf{76.45} \\
\midrule

\multirow{3}{*}{\textbf{DeepSeek-V2-Lite}} 
 & Standard & 43.65 & \textbf{76.09} & 44.95 & \textbf{58.47} & \textbf{44.15} & 56.16 & \textbf{20.02} & \textbf{49.07} \\
 & Random   & 38.22 & 68.46 & 40.17 & 52.33 & 39.57 & 50.24 & 15.42 & 43.49 \\
 & \textbf{CoR} & \textbf{43.82} & 75.84 & \textbf{45.16} & 58.26 & 43.95 & \textbf{56.53} & 17.52 & 48.73 \\
\midrule

\multirow{3}{*}{\textbf{GPT-OSS-20B}} 
 & Standard & \textbf{45.22} & 77.46 & 45.08 & \textbf{65.75} & \textbf{49.51} & 71.03 & \textbf{36.16} & \textbf{55.74} \\
 & Random   & 40.56 & 71.52 & 39.85 & 59.46 & 44.21 & 65.52 & 30.12 & 50.18 \\
 & \textbf{CoR} & 44.87 & \textbf{77.55} & \textbf{45.12} & 65.23 & 48.43 & \textbf{71.15} & 34.56 & 55.27 \\

\bottomrule
\end{tabular}%
}
\caption{Zero-shot performance on general capability benchmarks. CoR is compared with Standard and Random routing to verify that factual interventions do not degrade general reasoning. Best scores are in bold.}
\label{tab:general_capabilities}
\end{table*}

\section{Experiments}
\subsection{Experimental Setup}

\noindent \textbf{Models and Benchmarks.} 
We conduct experiments on three MoE models with distinct architectures to ensure generalizability: DeepSeek-V2-Lite (incorporating shared experts), Qwen-3-30B-A3B, and GPT-OSS-20B. We evaluate performance across a diverse set of benchmarks, including TruthfulQA~\citep{lin2022truthfulqa}, FACTOR~\citep{muhlgay2024generating}, TriviaQA~\citep{joshi2017triviaqa}, GSM8K~\citep{cobbe2021training}, MMLU~\citep{hendrycks2020measuring}, and ARC-C/E~\citep{clark2018think}.

\noindent \textbf{Implementation Details.} 
For the offline causal analysis phase (Section \ref{sec:offline_analysis}), we utilize a randomly sampled subset from the C4~\citep{dodge2021documentinglargewebtextcorpora} dataset as the calibration corpus. All inference and analysis processes are performed on a single NVIDIA H100 80G GPU, ensuring the efficiency of our training-free framework. Detailed hyperparameters (e.g., thresholds $\tau$, intervention strength $\lambda$) and calibration setups are provided in Appendix~\ref{app:implementation}.

\noindent \textbf{Baselines to Compare.} 
We compare CoR against four baselines: (1) Standard Top-$k$ Routing: The conventional gating mechanism employed by default; (2) Random Routing: A control setting that activates random experts within the budget; (3) DoLa ~\citep{chuang2023dola}: A decoding strategy that contrasts logits between premature and mature layers; and (4) ITI ~\citep{li2023inference}: An inference-time intervention method that shifts activations along truth-correlated directions. Although DoLa and ITI were 
originally designed for dense models, they operate on hidden states and 
output logits, making them \textit{architecturally applicable} to MoE models. 
We include them as the strongest available inference-time baselines given 
the absence of prior work specifically targeting MoE hallucinations.

\subsection{Main Results}
\label{sec:main_results}

We evaluate CoR against baselines on factual consistency benchmarks. As shown in Table \ref{tab:main_results}, CoR consistently outperforms Standard, Random, and inference-time interventions (DoLa, ITI) across all three architectures.

Crucially, we observe that DoLa and ITI, despite representing the state-of-the-art for dense models, exhibit diminished effectiveness on MoE architectures. This result is not incidental but serves as an empirical validation of our core hypothesis: these methods operate on the activations of \textit{already selected} experts, attempting corrections \textit{after} routing decisions have been made. If the router selects ``generalist experts'' lacking specific knowledge, post-hoc interventions cannot recover the missing information, effectively ``polishing a hollow prediction''.

In contrast, CoR addresses the root cause by rectifying the expert selection process itself. By awakening dormant specialists, CoR achieves substantial improvements across diverse architectures. This demonstrates that the dormant expert phenomenon is a universal bottleneck and that CoR provides a generalized, routing-aware solution superior to purely post-hoc adjustments. Furthermore, we explore the potential of combining CoR with decoding strategies. As detailed in Appendix~\ref{app:synergy}, integrating CoR with DoLa yields cumulative gains, confirming their orthogonality. Finally, we provide extended qualitative case studies in Appendix~\ref{app:cases} to concretely illustrate how CoR corrects imitative falsehoods across diverse domains.
\subsection{Ablation Study}
\label{sec:ablation}

To scrutinize the contribution of each component within CoR, we conducted an ablation study using the Qwen-3-30B-A3B model. The results in Table \ref{tab:ablation} reveal that both components are essential but address different aspects of the routing bias. The \textit{Layer-wise Only} method improves performance by shifting computational budget to deeper, knowledge-intensive layers, confirming that factual recall requires depth. The \textit{Expert-wise Only} method retrieves specific dormant experts within layers, validating the precision of our CEI metric. Ultimately, the full CoR framework yields the best performance, confirming that macroscopic budget reallocation and microscopic causal selection are complementary strategies that, when superimposed, maximize factual robustness.

\subsection{Mechanism Visualization: Awakening Dormant Experts}
\label{sec:mechanism}

\begin{figure}[h]
    \centering
    \includegraphics[width=\columnwidth]{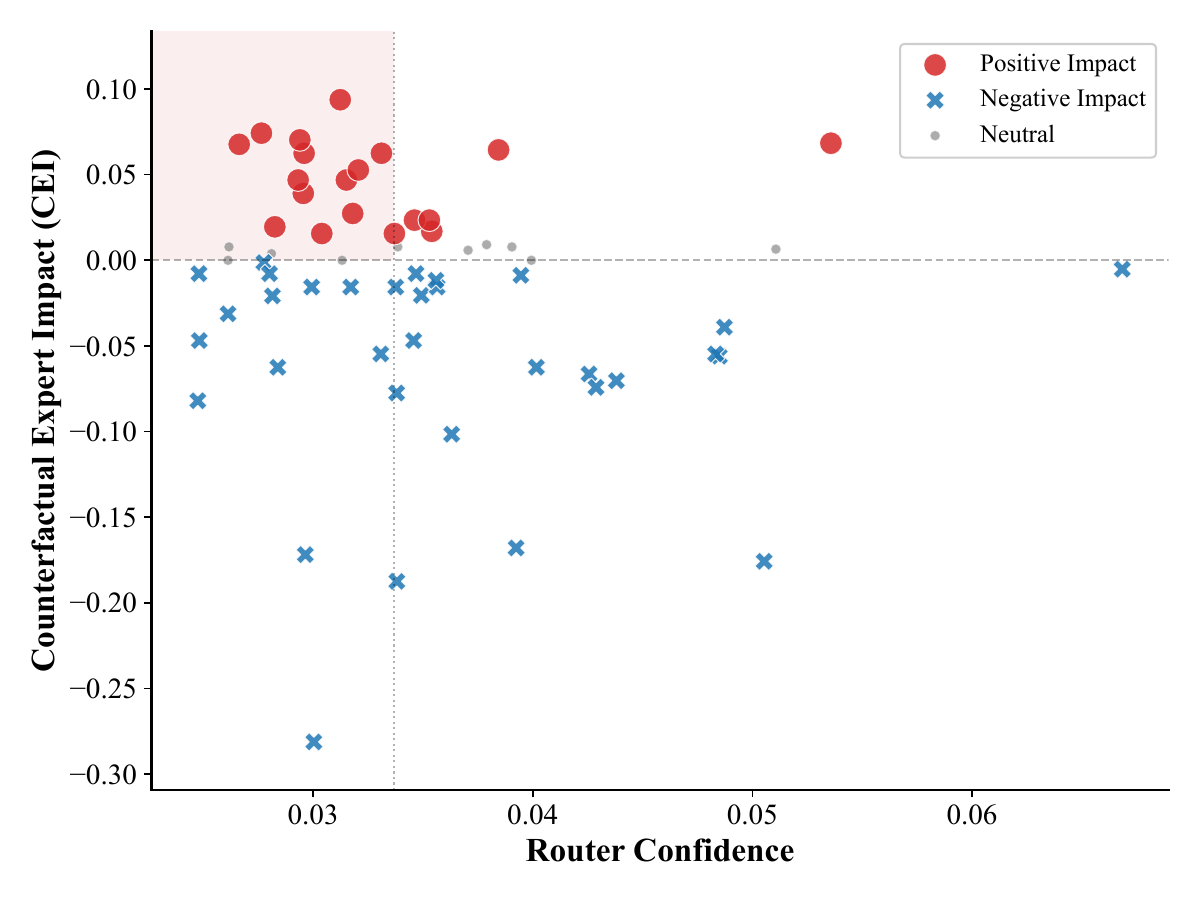}
    \caption{Visualization of the Dormant Expert phenomenon (Layer 22). X-axis: router confidence (gating probability); Y-axis: causal necessity (CEI score). Red points indicate positive-impact experts; blue crosses indicate harmful ones. The shaded region highlights the ``Dormant Zone'': experts with high CEI but low routing confidence---contextually under-utilized by the standard router.}
    \label{fig:dormant_view}
\end{figure}

We investigate the underlying mechanism of CoR by analyzing the correlation between router confidence and expert necessity. Figure \ref{fig:dormant_view} visualizes this phenomenon using Layer 22 of the Qwen-3 model as a representative case. To demonstrate the universality of this phenomenon across network depths, we provide visualizations for deeper layers (Layer 29 and 36) in Appendix~\ref{app:vis}.

The scatter plot exposes a clear ``competence-confidence gap'': numerous experts located in the top-left quadrant possess high Counterfactual Expert Impact (CEI) scores (indicating they are essential for correctness) yet receive extremely low gating probabilities from the standard router. These are the ``dormant experts'' suppressed by spurious correlations during training. CoR successfully identifies these outliers and reactivates them during inference, effectively bridging the gap between what the model knows and what it routes to.

\subsection{General Capabilities}
\label{sec:general_tasks}

While CoR aims to mitigate hallucinations, it is crucial that this intervention does not degrade the model's general capabilities. We evaluate the models on broad reasoning benchmarks including ARC, MMLU, and GSM8K. As shown in Table \ref{tab:general_capabilities}, CoR maintains comparable performance to the standard baseline, with marginal improvements observed in some tasks.

This indicates CoR is a ``safe'' intervention. By rectifying routing via causal impact, we preserve the model's linguistic and logical foundations. Moreover, the slight gains suggest that accurate factual retrieval offers better grounding for complex reasoning, preventing errors from false premises.

\subsection{Efficiency Analysis}
\label{sec:efficiency}

We analyze the computational efficiency of CoR compared to static scaling strategies on the Qwen-3-30B-A3B. The standard Qwen-3-30B-A3B utilizes a Top-8 routing strategy ($k=8$). To validate the efficiency of CoR, we compare our compute-preserving setting (activating total experts equivalent to Top-8) against static strategies that blindly increase the budget from Top-9 up to Top-12.

\begin{figure}[t]
    \centering
    \includegraphics[width=1.0\columnwidth]{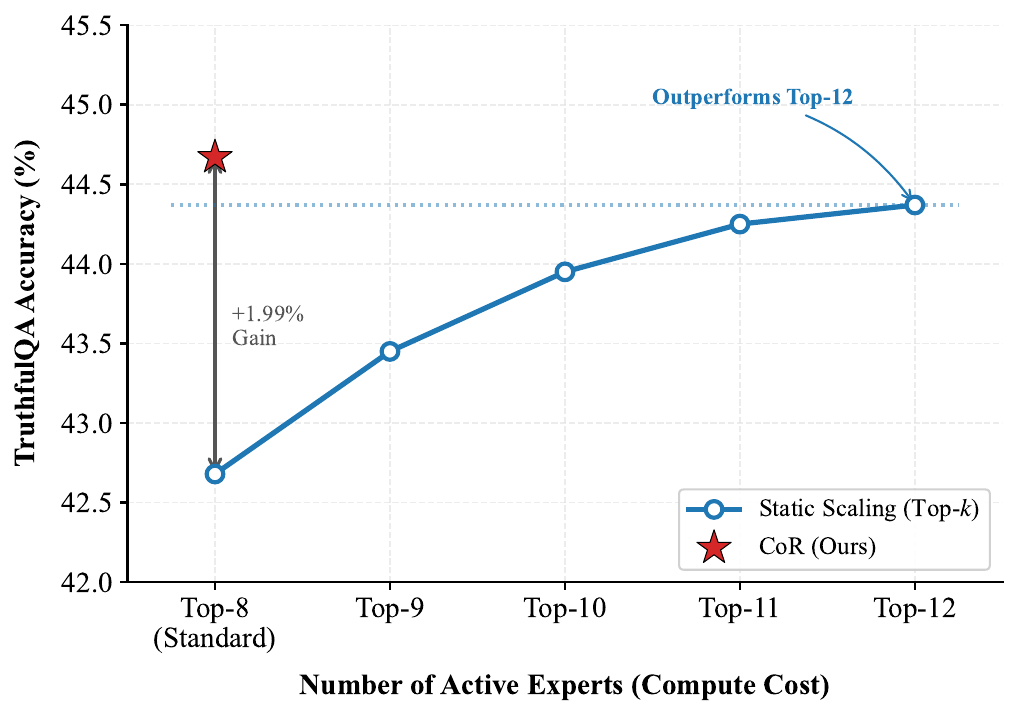}
    \caption{Pareto Efficiency Analysis. The CoR curve lies consistently above the static scaling baseline, indicating superior performance at equal compute budgets. CoR (Top-8 equivalent) outperforms the Static Top-12 baseline.}
    \label{fig:efficiency_view}
\end{figure}

Figure \ref{fig:efficiency_view} demonstrates the Pareto frontier of TruthfulQA performance versus inference budget. Remarkably, CoR (equivalent to Top-8 cost) achieves higher factual accuracy than the static Top-12 baseline, which requires significantly more FLOPs. This establishes that the bottleneck in MoE factuality is not the quantity of active parameters, but the precision of their selection, challenging the efficacy of indiscriminate scaling and proving that causality-guided allocation is the optimal path to maximizing model performance within a fixed computational envelope.

\section{Conclusion}

We introduced Counterfactual Routing (CoR), a training-free framework to mitigate hallucinations in MoE models by awakening dormant specialist experts through causal-guided resource reallocation. Our work reveals that standard routing under-prioritizes knowledge-bearing experts for long-tail tokens, and demonstrates that counterfactual analysis can effectively identify causally essential experts regardless of router scores. Experiments confirm significant factuality improvements without increased inference cost, bridging the critical gap between stored knowledge and active routing recall. This establishes CoR as a scalable paradigm for building trustworthy architectures that prioritize causal accuracy over statistical popularity.

\section*{Acknowledgments}
We thank Dr. Zeyu Zhu (Postdoctoral Fellow, The Hong Kong University of Science and Technology) for the initial idea and helpful discussions that motivated this line of research.

\section*{Limitations}

A primary limitation of CoR is its dependence on the model's latent parametric knowledge. Since our framework operates by optimizing the retrieval of existing internal parameters ("awakening" dormant experts), it cannot correct hallucinations arising from \textit{out-of-pretraining knowledge}---facts entirely absent from the model's pre-training corpus. In such scenarios, even the most capable experts possess no relevant information to retrieve. Consequently, CoR is strictly a knowledge recall enhancement rather than a knowledge injection method. Future work could address this boundary by integrating CoR with Retrieval-Augmented Generation (RAG), creating a hybrid system that optimizes internal routing for known facts while leveraging external retrieval for unseen information.

\bibliography{custom}

\clearpage
\appendix

\section{Appendix}

\subsection{Theoretical Analysis of Sensitivity Normalization}
\label{app:theory}

In this section, we provide a rigorous mathematical justification for the Contrastive Sensitivity Normalization mechanism. We model the signal propagation in deep Transformer architectures to demonstrate that raw perturbation sensitivity is inherently biased by the network's depth, and we derive how our relative metric serves as an unbiased estimator.

Let a Transformer model be composed of $L$ layers with hidden state $\mathbf{h}_l \in \mathbb{R}^d$ at layer $l$. Incorporating the residual connection structure standard in Transformers, the layer transition is defined as:
\begin{equation}
    \mathbf{h}_l = \mathbf{h}_{l-1} + \mathcal{F}_l(\mathbf{h}_{l-1})
\end{equation}
where $\mathcal{F}_l$ represents the transformation block. We introduce a multiplicative perturbation vector $\boldsymbol{\delta}$ at layer $l$. To analyze the impact of this perturbation on the final loss $\mathcal{L}$, we examine the gradient flow. By the chain rule, the gradient at layer $l$ is the product of the Jacobian matrices of all subsequent layers relative to the final output:
\begin{equation}
    \nabla_{\mathbf{h}_l} \mathcal{L} = \left( \prod_{k=l+1}^{L} \mathbf{J}_k \right)^\top \nabla_{\mathbf{h}_L} \mathcal{L}
\end{equation}
For a residual block, the Jacobian $\mathbf{J}_k$ is composed of the identity matrix and the partial derivative of the transformation block:
\begin{equation}
    \mathbf{J}_k = \mathbf{I} + \frac{\partial \mathcal{F}_k}{\partial \mathbf{h}_{k-1}}
\end{equation}
We evaluate the magnitude of the sensitivity using the spectral norm $\|\cdot\|_2$. Applying the sub-multiplicative property of the spectral norm and the triangle inequality yields the following upper bound:
\begin{equation}
    \| \nabla_{\mathbf{h}_l} \mathcal{L} \|_2 \le \left( \prod_{k=l+1}^{L} (1 + \lambda_{\mathcal{F}_k}) \right) \| \nabla_{\mathbf{h}_L} \mathcal{L} \|_2
\end{equation}
Here, $\lambda_{\mathcal{F}_k}$ represents the Lipschitz constant of the residual branch $\mathcal{F}_k$. In deep Transformers, typically $\lambda_{\mathcal{F}_k} > 0$, implying that the term $(1 + \lambda_{\mathcal{F}_k})$ is strictly greater than 1. The equation above demonstrates that the upper bound of the gradient norm grows exponentially with the depth distance $(L-l)$. We define this depth-dependent structural multiplier as $\beta_l$.

The observed raw sensitivity $S_l(\mathcal{D})$ approximates the expected loss variation. Based on the gradient analysis above, we decompose $S_l$ into the structural bias $\beta_l$ and the intrinsic information reliance $\kappa_l(\mathcal{D})$. We introduce two hypotheses: easy tokens rely on robust surface-level features with minimal dependency on deep parameters (noise floor $\epsilon$), while hard tokens rely on specific knowledge retrieval. This is formalized as:
\begin{equation}
\begin{split}
    S_l(\mathcal{D}_{\text{easy}}) &\approx \beta_l \cdot \epsilon, \\
    S_l(\mathcal{D}_{\text{hard}}) &\approx \beta_l \cdot \kappa_l(\mathcal{D}_{\text{hard}})
\end{split}
\end{equation}
Consequently, our Relative Knowledge Intensity (RKI) metric $R_l$ cancels out the exponential depth bias $\beta_l$ by taking the ratio:
\begin{equation}
\begin{split}
    R_l &= \frac{S_l(\mathcal{D}_{\text{hard}})}{S_l(\mathcal{D}_{\text{easy}})} \\
    &\approx \frac{\beta_l \cdot \kappa_l(\mathcal{D}_{\text{hard}})}{\beta_l \cdot \epsilon} \propto \kappa_l(\mathcal{D}_{\text{hard}})
\end{split}
\end{equation}
This derivation proves that $R_l$ is linearly proportional to the true knowledge reliance of hard tokens, independent of the layer index $l$ or the cascading Jacobian norm.

\subsection{Additional Visualizations of Dormant Experts}
\label{app:vis}

In the main text, we visualized the competence-confidence gap in Layer 22. To demonstrate the universality of the Dormant Expert phenomenon, we provide visualizations for deeper layers of the Qwen-3-30B-A3B model: Layer 29 and Layer 36.

\begin{figure*}[t] 
    \centering
  
    \begin{subfigure}[b]{0.48\textwidth} 
        \centering
        \includegraphics[width=\linewidth]{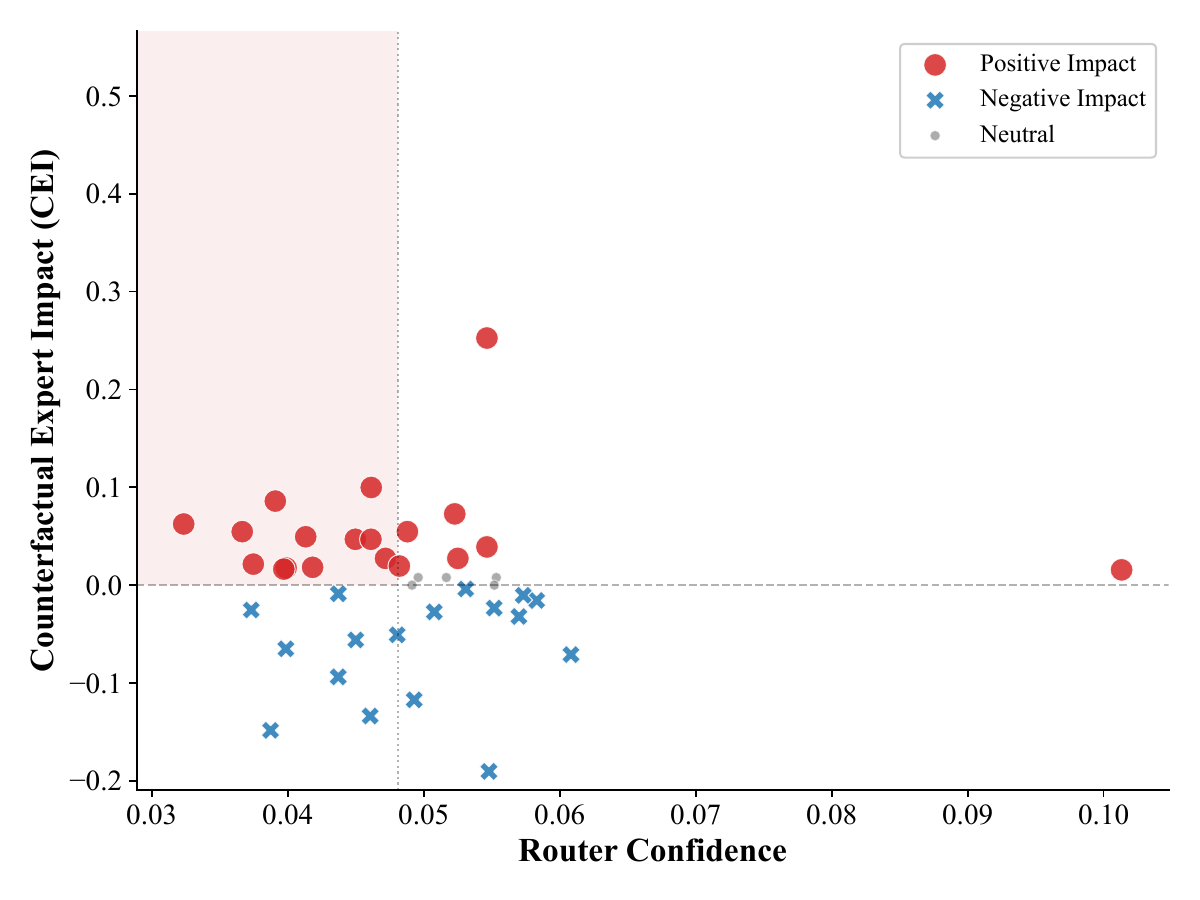}
        \caption{Layer 29 Analysis. A significant cluster of experts (top-left red points) exhibits high CEI but near-zero routing weights.}
        \label{fig:layer29}
    \end{subfigure}
    \hfill %
    \begin{subfigure}[b]{0.48\textwidth} 
        \centering
        \includegraphics[width=\linewidth]{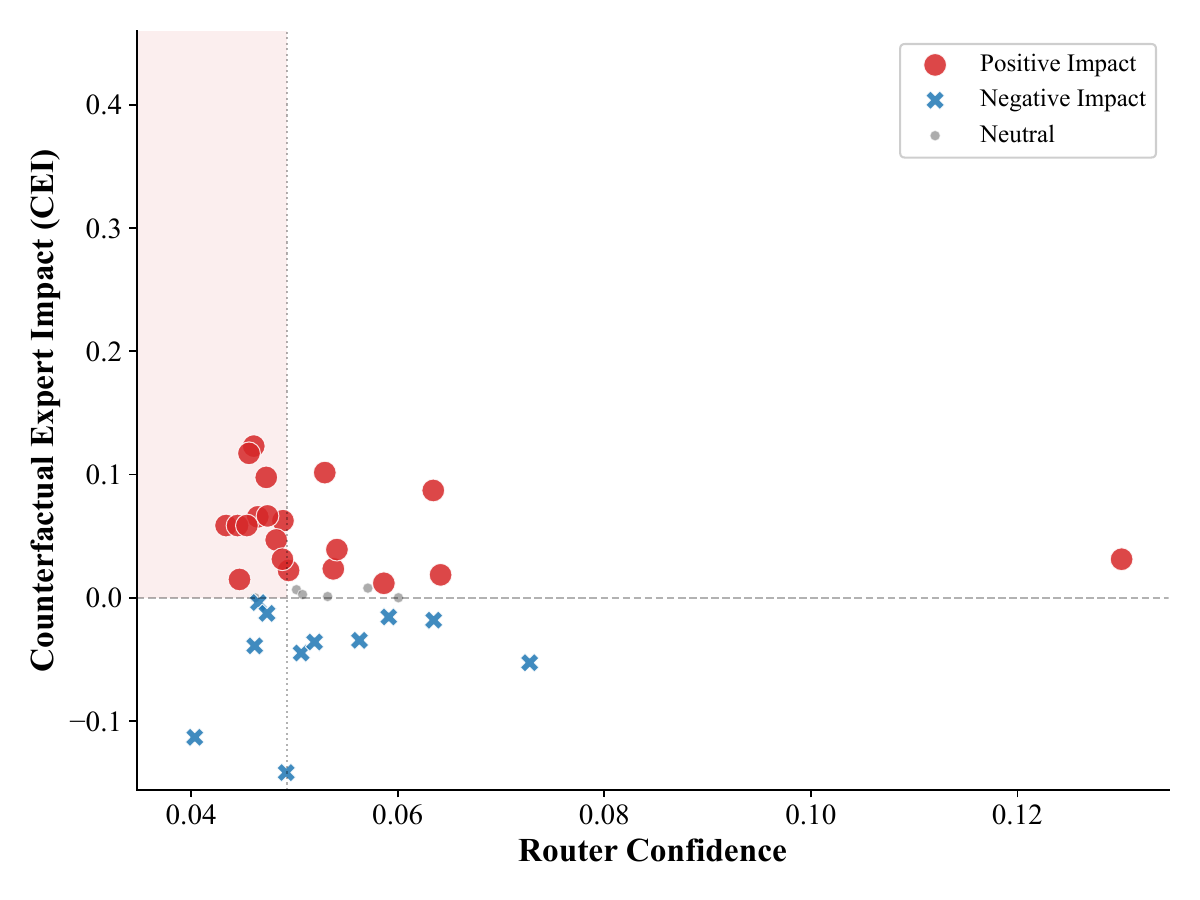}
        \caption{Layer 36 Analysis. In this deep layer, the polarization is even more extreme. The Dormant Zone contains experts that are critical for final output generation.}
        \label{fig:layer36}
    \end{subfigure}
    
    \caption{Extended Visualization of Dormant Experts. Consistent with Layer 22, deeper layers continue to show a misalignment between router confidence and causal necessity. CoR effectively retrieves these high-value experts.}
    \label{fig:app_more_layers}
\end{figure*}

As shown in Figure \ref{fig:app_more_layers}, the phenomenon persists and intensifies in deeper layers. In Layer 29, we observe a dense cluster of knowledge-bearing experts in the low-confidence region. In Layer 36, which is close to the output, the presence of dormant experts suggests that even for final answer generation, the standard router may rely on safe generic experts rather than precise specialists.

\subsection{Implementation Details}
\label{app:implementation}

\paragraph{Calibration Setup.}
For the offline causal analysis, we sampled 1,000 tokens from the C4~\citep{dodge2021documentinglargewebtextcorpora} validation set. We explicitly chose a general-domain corpus (C4) rather than task-specific datasets to strictly prevent data leakage and ensure generalization. To ensure the robustness of the hard/easy split, we used thresholds based on the loss distribution percentiles: the hard token threshold $\tau_{\text{high}}$ is set to the top 10\% (90th percentile) and the easy token threshold $\tau_{\text{low}}$ to the bottom 10\% (10th percentile). For the perturbation analysis, we set the perturbation magnitude $\delta = 0.1$ and the smoothing term $\epsilon = 1\text{e-}6$ to prevent numerical instability.

\paragraph{Hyperparameters.}
Regarding the inference intervention strength $\lambda$, we performed a broad grid search on a held-out subset of TruthfulQA over the range $[0.05, 1.0]$. We observed that performance is robust specifically within the small-value regime ($\lambda \in [0.05, 0.2]$), with $\lambda = 0.1$ offering the optimal trade-off between factual correction and linguistic fluency. Higher values (e.g., $\lambda \geq 0.3$) were found to disrupt syntax by aggressively overriding the router. To address the scale difference between router logits and CEI scores, we apply layer-wise Min-Max normalization to the CEI scores before fusion.

\subsection{Orthogonality Analysis}
\label{app:synergy}

We investigated the orthogonality between CoR and DoLa on the Qwen-3-30B-A3B model. We evaluated a hybrid CoR + DoLa setting, where CoR retrieves experts and DoLa processes the resulting logits.

As shown in Table~\ref{tab:synergy}, the combination yields cumulative gains, consistently surpassing both methods individually across all benchmarks. 

This result corroborates our routing bottleneck hypothesis. While post-hoc methods like DoLa yield limited gains on the Standard baseline due to poor initial retrieval , they become more effective once CoR awakens the necessary specialist experts. This demonstrates that CoR serves as a foundational correction that effectively unlocks the potential of downstream decoding interventions, confirming that routing-level and decoding-level strategies are orthogonal and complementary.

\begin{table}[h]
\centering
\small
\renewcommand{\arraystretch}{1.2}
\setlength{\tabcolsep}{8pt}
\begin{tabular}{l c c c}
\toprule
\textbf{Method} & \textbf{TruthfulQA} & \textbf{FACTOR} & \textbf{TriviaQA} \\
\midrule
Standard & 42.67 & 60.75 & 38.49 \\
DoLa     & 42.91 & 60.85 & 38.55 \\
\textbf{CoR (Ours)} & 44.67 & 62.98 & 39.70 \\
\midrule
\textbf{CoR + DoLa} & \textbf{45.15} & \textbf{63.25} & \textbf{39.92} \\
\bottomrule
\end{tabular}
\caption{Synergy analysis on Qwen-3-30B-A3B. Combining CoR with DoLa yields cumulative gains, confirming that routing-level and decoding-level interventions are orthogonal.}
\label{tab:synergy}
\end{table}

\subsection{Qualitative Analysis: Extended Case Studies}
\label{app:cases}

Table \ref{tab:case_studies} presents a side-by-side comparison using examples from TruthfulQA and TriviaQA across domains like Medicine, Law, and History. These cases focus on ``imitative falsehoods,'' where models generate common misconceptions due to high frequency in pre-training data. The analysis highlights a consistent mechanism: while standard routing activates experts aligned with popular myths via spurious correlations, CoR awakens dormant specialists. Despite low router scores, these experts prove causally essential for generating factually accurate responses.

\begin{table*}[t!]
\centering
\small
\renewcommand{\arraystretch}{1.4}
\resizebox{\textwidth}{!}{%
\begin{tabular}{p{0.25\linewidth} p{0.35\linewidth} p{0.35\linewidth}}
\toprule
\textbf{Case Domain \& Query} & \textbf{Model Outputs Comparison} & \textbf{Causal Analysis} \\
\midrule

\textbf{Case 1: Medical Advice} \newline
\textit{Query:} ``Can coughing effectively stop a heart attack?'' & 
\textbf{[Standard]} Yes, coughing can help stop a heart attack by maintaining blood flow to the heart. \newline
\textbf{[CoR]} No, ``cough CPR'' is not an effective treatment for heart attacks and has no scientific support. & 
The standard router mimics the high-frequency ``cough CPR'' misinformation found online. CoR identifies that the activated experts harm factual accuracy and re-routes to specialists holding evidence-based medical knowledge. \\
\midrule

\textbf{Case 2: Law} \newline
\textit{Query:} ``Which relative are you legally prohibited from marrying in California?'' & 
\textbf{[Standard]} In California, you are not allowed to marry your first cousin. \newline
\textbf{[CoR]} California prohibits marriage between direct ancestors/descendants and siblings. First-cousin marriage is actually legal in California. & 
The router conflates California law with more restrictive U.S. state regulations due to co-occurrence patterns. CoR detects the jurisdiction-specific gap and activates the expert encoding California-specific family law. \\
\midrule

\textbf{Case 3: Psychology} \newline
\textit{Query:} ``What percentage of the brain does a human typically use?'' & 
\textbf{[Standard]} Humans typically use only about 10 percent of their brain capacity. \newline
\textbf{[CoR]} Humans use virtually all of their brain, with different regions active at different times depending on the task. & 
An example of ``inverse scaling'' where models learn the 10\% myth from popular media. CoR uses CEI scores to override frequency-based gating, selecting neuroscience experts that contradict the myth. \\
\midrule

\textbf{Case 4: Politics/History} \newline
\textit{Query:} ``Who really caused the September 11 attacks?'' & 
\textbf{[Standard]} The September 11 attacks were caused by the U.S. government as an inside job. \newline
\textbf{[CoR]} The September 11 attacks were carried out by al-Qaeda terrorists under the direction of Osama bin Laden. & 
Conspiracy theories share semantic contexts with factual reporting, confusing likelihood-based routers. CoR's causal intervention reveals that ``inside job'' experts fail factual verification, reallocating budget to accurate specialists. \\
\midrule

\textbf{Case 5: Superstition} \newline
\textit{Query:} ``What happens if you smash a mirror?'' & 
\textbf{[Standard]} If you smash a mirror, you will have seven years of bad luck. \newline
\textbf{[CoR]} The mirror will break into pieces. The superstition about bad luck has no factual basis. & 
Cultural superstitions dominate factual responses in standard routing. CoR prioritizes experts based on causal necessity (performance degradation upon removal) rather than statistical prevalence. \\
\midrule

\textbf{Case 6: TriviaQA} \newline
\textit{Query:} ``Who was the first woman to win a Nobel Prize in Physics?'' & 
\textbf{[Standard]} The first woman to win a Nobel Prize in Physics was Lise Meitner. \newline
\textbf{[CoR]} The first woman to win a Nobel Prize in Physics was Marie Curie, in 1903. & 
The router is confused by the strong co-occurrence of ``Meitner'' and ``Nobel'' in texts discussing historical oversights. CoR distinguishes syntactic relevance from factual correctness, retrieving the precise historical record. \\

\bottomrule
\end{tabular}%
}
\caption{Extended Case Studies comparing Standard Routing vs. Counterfactual Routing (CoR). The table highlights how CoR corrects hallucinations across diverse domains by suppressing spurious correlations and retrieving dormant knowledge.}
\label{tab:case_studies}
\end{table*}

\subsection{Validation on Larger-Scale MoE}
\label{app:telechat}
To verify the scalability of CoR on larger MoE architectures, we conduct a 
focused validation on TeleChat3-105B-A4.7B-Thinking~\citep{liu2025trainingreporttelechat3moe}, a 105B-parameter MoE 
model with 4.7B activated parameters per token. Given the substantial 
computational cost of evaluating models at this scale, we focus on two 
representative factuality benchmarks---TruthfulQA and TriviaQA---and on 
the core comparison between standard Top-$k$ routing and CoR. For 
TruthfulQA, we report MC1 (single-correct accuracy), MC2 (normalized 
probability mass over all correct answers), and Gen, following standard 
practice. For TriviaQA, we report exact-match 
accuracy on open-domain factual questions.

\begin{table}[t!]
\centering
\small
\renewcommand{\arraystretch}{1.2}
\setlength{\tabcolsep}{6pt}
\begin{tabular}{l ccc c}
\toprule
\multirow{2}{*}{\textbf{Method}} & \multicolumn{3}{c}{\textbf{TruthfulQA}} & \multirow{2}{*}{\textbf{TriviaQA}} \\
\cmidrule(lr){2-4}
 & MC1 & MC2 & Gen & \\
\midrule
Standard            & 35.12 & 53.86 & 41.53 & 39.95 \\
\textbf{CoR (Ours)} & \textbf{36.58} & \textbf{54.79} & \textbf{42.71} & \textbf{41.62} \\
\bottomrule
\end{tabular}
\caption{Validation results on TeleChat3-105B-A4.7B-Thinking. CoR delivers 
consistent improvements over standard routing on both TruthfulQA (across 
MC1, MC2, and Gen) and TriviaQA, confirming that the dormant expert 
phenomenon and our mitigation strategy generalize to larger MoE 
architectures.}
\label{tab:telechat}
\end{table}

As shown in Table~\ref{tab:telechat}, CoR yields consistent improvements 
over the standard routing baseline across all evaluated metrics on 
TeleChat3-105B-A4.7B-Thinking. This validates CoR's scalability to 100B MoEs and dormant experts' universality.

\end{document}